\documentclass{article}

\usepackage{microtype}
\usepackage{graphicx}
\usepackage{subcaption}
\usepackage{booktabs} 
\usepackage{amsfonts}
\usepackage{amsmath}

\usepackage{hyperref}

\usepackage[accepted]{icml2019}

\icmltitlerunning{Distributional Reinforcement Learning for Efficient Exploration}

\begin{document}

\twocolumn[
\icmltitle{Distributional Reinforcement Learning for Efficient Exploration}

\begin{icmlauthorlist}
\icmlauthor{Borislav Mavrin}{to,hu}
\icmlauthor{Shangtong Zhang}{ox}
\icmlauthor{Hengshuai Yao}{si}
\icmlauthor{Linglong Kong}{to,hu}
\icmlauthor{Kaiwen Wu}{uw}
\icmlauthor{Yaoliang Yu}{uw}
\end{icmlauthorlist}

\icmlaffiliation{to}{University of Alberta}
\icmlaffiliation{hu}{Huawei Noah's Ark}
\icmlaffiliation{si}{Huawei Hi-Silicon}
\icmlaffiliation{ox}{University of Oxford. Work done during an internship with Huawei.}
\icmlaffiliation{uw}{University of Waterloo}


%

\icmlcorrespondingauthor{Hengshuai Yao}{hengshuai.yao@huawei.com}

\icmlkeywords{Deep Reinforcement Learning, Exploration}

\vskip 0.3in
]



\printAffiliationsAndNotice{}  


\begin{abstract}  
In distributional reinforcement learning (RL), the estimated distribution of value function models both the parametric and intrinsic uncertainties.
We propose a novel and efficient exploration method for deep RL that has two components.  
The first is a decaying schedule to suppress the intrinsic uncertainty. The second is an exploration bonus calculated from the upper quantiles of the learned distribution. 
In Atari 2600 games, our method outperforms QR-DQN in 12 out of 14 hard games (achieving 483 \% average gain across 49 games in cumulative rewards over QR-DQN with a big win in Venture).  
We also compared our algorithm with QR-DQN in a challenging 3D driving simulator (CARLA). 
Results show that our algorithm achieves near-optimal safety rewards twice faster than QRDQN. 
\end{abstract}

\section{Introduction}
Exploration is a long standing problem in Reinforcement Learning (RL), where \textit{optimism in the face of uncertainty} is one fundamental principle \cite{lai1985asymptotically,strehl2005theoretical}. Here the uncertainty refers to \textit{parametric uncertainty}, which arises from the variance in the estimates of certain parameters given finite samples. Both count-based methods \cite{auer2002using,kaufmann2012bayesian,bellemare2016unifying,ostrovski2017count,tang2017exploration} and Bayesian methods \cite{kaufmann2012bayesian,chen2017ucb,o2017uncertainty} follow this optimism principle. In this paper, we propose to use distributional RL methods to achieve this optimism.

Different from classical RL methods, where an expectation of value function is learned \cite{sutton1988learning,watkins1992q,mnih2015human}, distributional RL methods \cite{jaquette1973markov,bellemare2017distributional} maintain a full distribution of future return. In the limit, distributional RL captures the intrinsic uncertainty of an MDP \cite{bellemare2017distributional,dabney2017distributional,dabney2018implicit,rowland2018analysis}. \textit{Intrinsic uncertainty arises from the stochasticity of the environment}, which is parameter and sample independent. However, it is not trivial to quantify the effects of parametric and intrinsic uncertainties in distribution learning. To investigate this, let us look closer at a simple setup of distribution learning. Here we use Quantile Regression (QR) (detailed in Section 2.2), but the example presented here holds for other distribution learning methods. Here the random samples are drawn from any stationary distribution. The initial estimated distribution is set to be the uniform one (left plots). At each time step, QR updates its estimate in an on-line fashion by minimizing some loss function. In the limit the estimated QR distribution converges to the true distribution (right plots). The two middle plots examine the intermediate estimated distributions before convergence in two distinct cases.

        \begin{figure}
        \centering
        \begin{subfigure}[b]{0.8\linewidth}
        \includegraphics[width=1\linewidth]{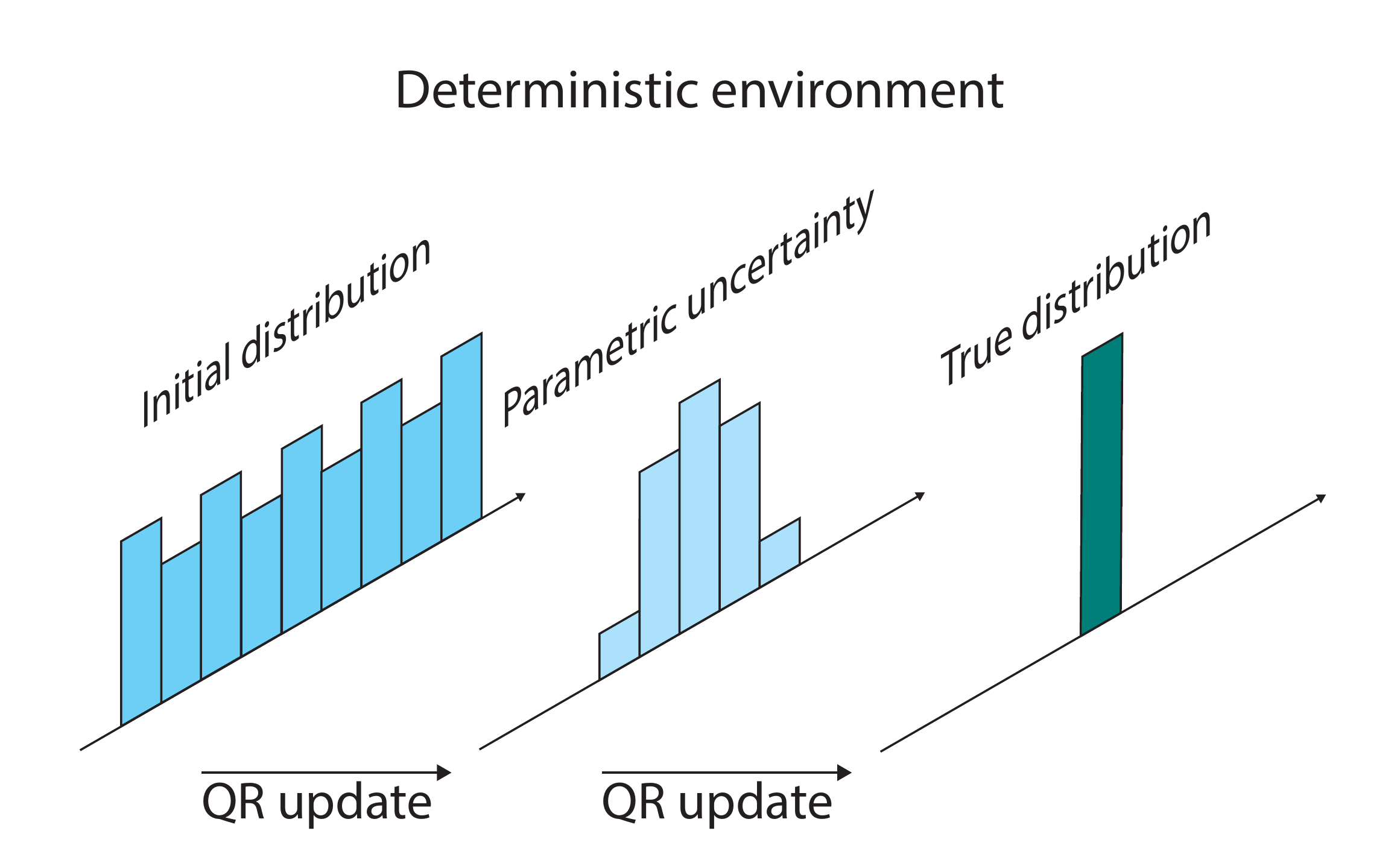}
            \caption{Intrinsic uncertainty.}
            \label{fig:qr_update_a}
        \end{subfigure}
        \begin{subfigure}[b]{0.8\linewidth}
        \includegraphics[width=1\linewidth]{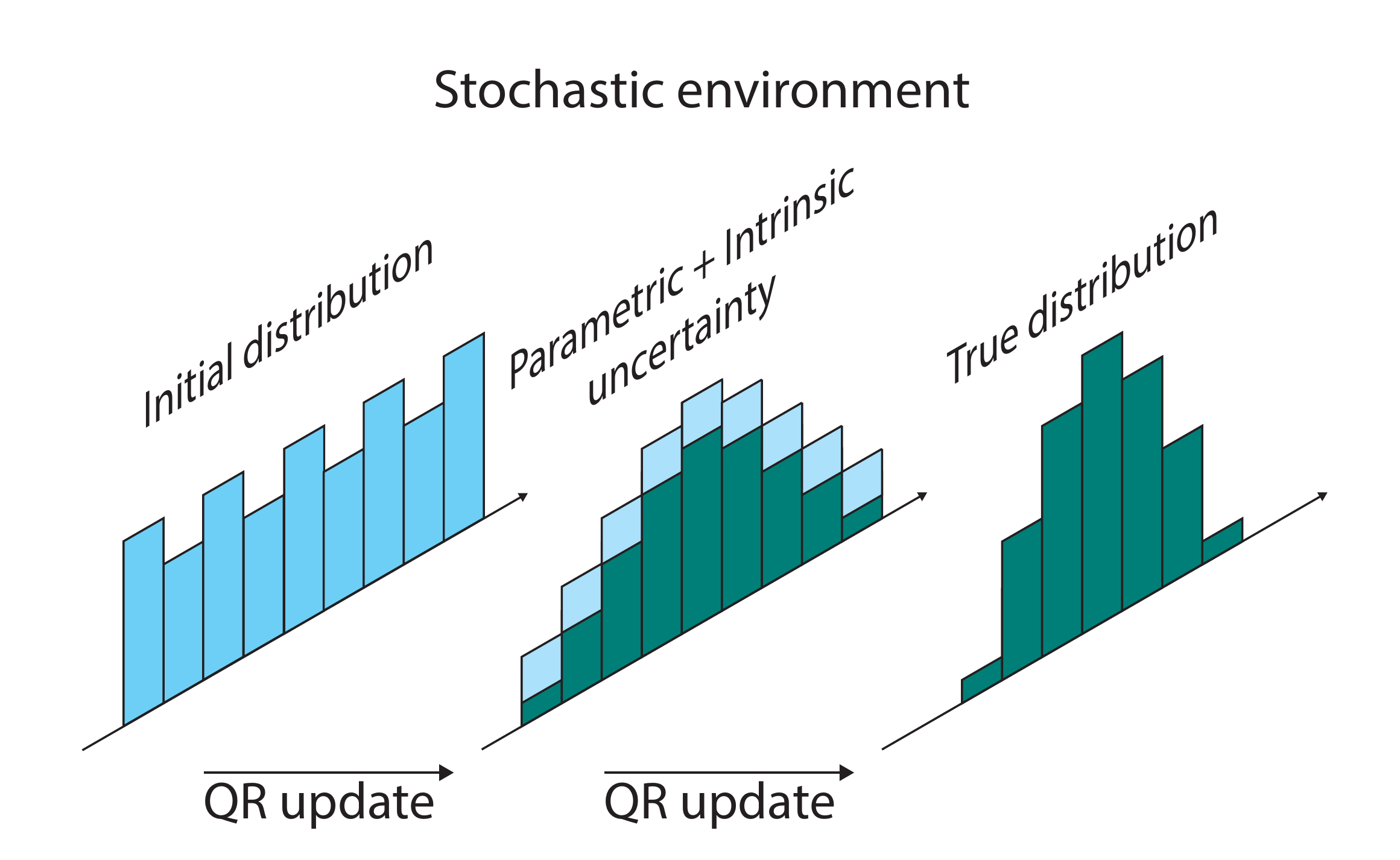}
            \caption{Intrinsic and parametric uncertainties.}
            \label{fig:qr_update_b}
        \end{subfigure}
        \caption{Uncertainties in deterministic and stochastic environments.}
    \end{figure}

Case 1: Figure \ref{fig:qr_update_a} shows a deterministic environment where the data is generated by a degenerate distribution. 
In this case, the intermediate estimate of the distribution (middle plot) contains only the information about parametric uncertainty. Here, parametric uncertainty comes from the error in the estimation of the quantiles. 
The left sub-plot shows estimation from the initialized parameters for the distribution estimator.
The middle sub-plot shows the estimated distribution converges closer to the true distribution on the right sub-plot.   

Case 2: Figure \ref{fig:qr_update_b} shows a stochastic environment, where the data is generated by a non-degenerate (stationary) distribution. In this case, the intermediate estimated distribution is the result of both parametric and intrinsic uncertainties. 
In the middle plot, the distribution estimator (QR) models randomness from both parametric and intrinsic uncertainties, and it is hard to split them. The parametric uncertainty does go away over time and converge to the true distribution on the right sub-plot.  
Our main insight in this paper is that the upper bound for a state-action value estimate shrinks at a certain rate (See Section \ref{sec:algorithm} for details). Specifically, the error of the quantile estimator is known to converge asymptotically in distribution to the Normal distribution \cite{koenker2005quantile}. By treating the estimated distribution during learning as sub-normal we can estimate the upper bound  of the state-action values with a high confidence (by applying Hoeffding’s inequality).

This example illustrates distributions learned via distributional methods (such as distributional RL algorithms) model the randomness arising from both intrinsic and parametric uncertainties. In this paper, we study how to take advantage of distributions learned by distributional RL methods for efficient exploration in the face of uncertainty. 

To be more specific, we use Quantile Regression Deep-Q-Network (QR-DQN, \cite{dabney2017distributional}) to learn the distribution of value function. 
We start with an examination of the two uncertainties and a naive solution that leaves the intrinsic uncertainty unsupressed. We construct a counter example in which this naive solution fails to learn. 
The intrinsic uncertainty persists and leads the naive solution to favor actions with higher variances. 
To suppress the intrinsic uncertainty, 
we apply a decaying schedule to improve the naive solution.

One interesting finding in our experiments is that the distributions learned by QR-DQN can be asymmetric. 
By using the upper quantiles of the estimated distribution  \cite{mullooly1988variance}, we estimate an optimistic exploration bonus for QR-DQN. 

We evaluated our algorithm in 49 Atari games \cite{bellemare2013arcade}. Our approach achieved 483 \% average gain in cumulative rewards over QR-DQN. The overall improvement is reported in Figure~\ref{fig:atari_results}.

We also compared our algorithm with QR-DQN in a challenging 3D driving simulator (CARLA). 
Results show that our algorithm achieves near-optimal safety rewards twice faster than QRDQN. 

In the rest of this paper, we first present some preliminaries of RL Section 2. In Section 3, we then study the challenges posed by the mixture of parametric and intrinsic uncertainties, and propose a solution to suppress the intrinsic uncertainty. We also propose a truncated variance estimation for exploration bonus in this section. In Section 4, we present empirical results in Atari games. Section 5 contains results on CARLA. Section 6 an overview of related work, and Section 7 contains conclusion.

\section{Background}
    \subsection{Reinforcement Learning}
        We consider a Markov Decision Process (MDP) of a state space $\mathcal{S}$, an action space $\mathcal{A}$, a reward ``function'' $R: \mathcal{S} \times \mathcal{A} \rightarrow \mathbb{R}$, a transition kernel $p: \mathcal{S} \times \mathcal{A} \times \mathcal{S} \rightarrow [0, 1]$, and a discount ratio $\gamma \in [0, 1)$. In this paper we treat the reward ``function'' $R$ as a random variable to emphasize its stochasticity. Bandit setting is a special case of the general RL setting, where we usually only have one state.
        
        We use $\pi: \mathcal{S} \times \mathcal{A} \rightarrow [0, 1]$ to denote a stochastic policy. We use $Z^\pi(s, a)$ to denote the random variable of the sum of the discounted rewards in the future, following the policy $\pi$ and starting from the state $s$ and the action $a$. We have $Z^\pi(s, a) \doteq \sum_{t=0}^\infty \gamma^t R(S_t, A_t)$, where $S_0 = s, A_0 = a$ and $S_{t+1} \sim p(\cdot | S_t, A_t), A_t \sim \pi(\cdot| S_t)$. The expectation of the random variable $Z^\pi(s, a)$ is $$Q^\pi(s, a) \doteq \mathbb{E}_{\pi, p, R}[Z^\pi(s, a)]$$ which is usually called the state-action value function. 
        In general RL setting, we are usually interested in finding an optimal policy $\pi^*$, such that $Q^{\pi^*}(s, a) \geq Q^\pi(s, a)$ holds for any $(\pi, s, a)$. All the possible optimal policies share the same optimal state-action value function $Q^*$, which is the unique fixed point of the Bellman optimality operator \cite{bellman2013dynamic},
        \begin{align*}
        Q(s, a) = \mathcal{T}Q(s, a) \doteq \mathbb{E}[R(s, a)] + \gamma \mathbb{E}_{s^\prime \sim p}[\max_{a^\prime} Q(s^\prime, a^\prime)]
        \end{align*}
        Based on the Bellman optimality operator, \citet{watkins1992q} proposed Q-learning to learn the optimal state-action value function $Q^*$ for control. At each time step, we update $Q(s,a)$ as 
        \begin{align*} 
        Q(s, a) \leftarrow Q(s, a) + \alpha (r + \gamma \max_{a^\prime}Q(s^\prime, a^\prime) - Q(s, a))
        \end{align*}
        where $\alpha$ is a step size and $(s, a, r, s^\prime)$ is a transition. There have been many work extending Q-learning to linear function approximation \cite{sutton2018reinforcement,szepesvari2010algorithms}. \citet{mnih2015human} combined Q-learning with deep neural network function approximators, resulting the Deep-Q-Network (DQN). Assume the $Q$ function is parameterized by a network $\theta$, at each time step, DQN performs a stochastic gradient descent to update $\theta$ minimizing the loss
        \begin{align*}
        \frac{1}{2}(r_{t+1} + \gamma \max_a Q_{\theta^-}(s_{t+1}, a) - Q_\theta(s_t, a_t)) ^ 2
        \end{align*}
        where $\theta^-$ is target network \cite{mnih2015human}, which is a copy of $\theta$ and is synchronized with $\theta$ periodically, and $(s_t, a_t, r_{t+1}, s_{t+1})$ is a transition sampled from a experience replay buffer \cite{mnih2015human}, which is a first-in-first-out queue storing previously experienced transitions. Decorrelating representation has shown to speed up DQN significantly \cite{mavrin_decor}. For simplicity, in this paper we will focus on the case without decorrelation. 
        
    \subsection{Quantile Regression}
        The core idea behind QR-DQN is the Quantile Regression introduced by the seminal paper \cite{koenker1978regression}. This approach gained significant attention in the field of Theoretical and Applied Statistics and might not be well known in other fields. For that reason we give a brief introduction here. Let us first consider QR in the supervised learning. Given data $\{(x_i, y_i)\}_i$, we want to compute the quantile of $y$ corresponding the quantile level $\tau$.  linear quantile regression loss is defined as:
        \begin{equation} \label{qr-loss}
            L(\beta) = \sum_i \rho_\tau(y_i - x_i \beta)
        \end{equation}
        where 
        \begin{equation} \label{check-function}
        \rho_\tau(u) = u (\tau - I_{u < 0}) = \tau |u| I_{u \ge 0} + (1 - \tau) |u| I_{u < 0}
        \end{equation}
        is the weighted sum of residuals. Weights are proportional to the counts of the residual signs and order of the estimated quantile $\tau$. For higher quantiles positive residuals get higher weight and vice versa.
        If $\tau=\frac{1}{2}$, then the estimate of the median for $y_i$ is $\theta_1 (y_i|x_i) = x_i \hat{\beta}$, with $\hat{\beta} = \arg \min L(\beta)$.
        
        

    \subsection{Distributional RL}
        Instead of learning the expected return $Q$, distributional RL focuses on learning the full distribution of the random variable $Z$ directly \cite{jaquette1973markov,bellemare2017distributional,mavrin_aamas_abstract}. There are various approaches to represent a distribution in RL setting \cite{bellemare2017distributional,dabney2018implicit,barth2018distributed}. In this paper, we focus on the quantile representation \cite{dabney2017distributional} used in QR-DQN, where the distribution of $Z$ is represented by a uniform mix of $N$ supporting quantiles:
        \begin{align*}
        Z_\theta(s, a) \doteq \frac{1}{N}\sum_{i=1}^N \delta_{\theta_i(s, a)}
        \end{align*}
        where $\delta_x$ denote a Dirac at $x \in \mathbb{R}$, and each $\theta_i$ is an estimation of the quantile corresponding to the quantile level (a.k.a. quantile index) $\hat{\tau}_i \doteq \frac{\tau_{i-1} + \tau_i}{2}$ with $\tau_i \doteq \frac{i}{N}$ for $0 \leq i \leq N$. The state-action value $Q(s, a)$ is then approximated by $\frac{1}{N}\sum_{i=1}^N \theta_i(s, a)$. Such approximation of a distribution is referred to as quantile approximation. 
        
        Similar to the Bellman optimality operator in mean-centered RL, we have the distributional Bellman optimality operator for control in distributional RL,
        \begin{align*}
        \mathcal{T}Z(s, a) \doteq R(s, a) + \gamma Z(s^\prime, \arg\max_{a^\prime}\mathbb{E}_{p, R}[Z(s^\prime, a^\prime)]) \\
        s^\prime \sim p(\cdot|s, a)
        \end{align*}
        Based on the distributional Bellman optimality operator, \citet{dabney2017distributional} proposed to train quantile estimations (i.e., $\{q_i\}$) via the Huber quantile regression loss \cite{huber1964robust}. To be more specific, at time step $t$ the loss is 
        \begin{align*}
        \frac{1}{N}\sum_{i=1}^N \sum_{i^\prime=1}^N\Big[\rho_{\hat{\tau}_i}^\kappa\big(y_{t, i^\prime} - \theta_i(s_t, a_t)\big)\Big]
        \end{align*}
        where $y_{t, i^\prime} \doteq r_t + \gamma \theta_{i^\prime}\big(s_{t+1}, \arg\max_{a^\prime}\sum_{i=1}^N \theta_i(s_{t+1}, a^\prime)\big)$ and 
        $\rho_{\hat{\tau}_i}^\kappa(x) $ \\
        $\doteq |\hat{\tau}_i - \mathbb{I}\{x < 0\}|\mathcal{L}_\kappa(x)$, where $\mathbb{I}$ is the indicator function and $\mathcal{L}_\kappa$ is the Huber loss,
        \begin{align*}
            \mathcal{L}_\kappa(x) \doteq \begin{cases}
            \frac{1}{2}x^2 & \text{if } x \leq \kappa \\
            \kappa(|x| - \frac{1}{2}\kappa) & \text{otherwise}
            \end{cases}
        \end{align*}

\section{Algorithm}\label{sec:algorithm}
    In this section we present our method. First, we study the issue of the mixture of parametric and intrinsic uncertainties in the estimated distributions learned by QR approach. We show that the intrinsic uncertainty has to be suppressed in calculating exploration bonus and introduce a decaying schedule to achieve this. 
    
    Second, in a simple example where the distribution is asymmetric, we show exploration bonus from truncated variance outperforms bonus from the variance. In fact, we did find that the distributions learned by QR-DQN (in Atari games) can be asymmetric. Thus we combine the truncated variance for exploration in our method. 

    \subsection{The issue of intrinsic uncertainty}
    
    A naive approach to exploration would be to use the variance of the estimated distribution as a bonus. We provide an illustrative counter example. Consider a multi-armed bandit environment with 10 arms where each arm's reward follows normal distribution $\mathcal{N}(\mu_k, \sigma_k)$. In each run, means $\{\mu_k\}_k$ are drawn from standard normal. Standard deviation of the best arm is set to 1.0, other arms' standard deviations are set to 5.
    In the setting of multi-armed bandits, this approach leads to picking the arm $a$ such that
    \begin{equation} \label{eq:naive_bandit}
        a = \arg \max_k \bar{\mu}_k + c \sigma_k
    \end{equation}
    where $\bar{\mu}_k$ and $\sigma^2_k$ are the estimated mean and variance of the $k$-th arm, computed from the corresponding quantile distribution estimation.
    
\begin{figure}
        \centering
        \begin{subfigure}{0.42\textwidth}
        \includegraphics[width=1\textwidth]{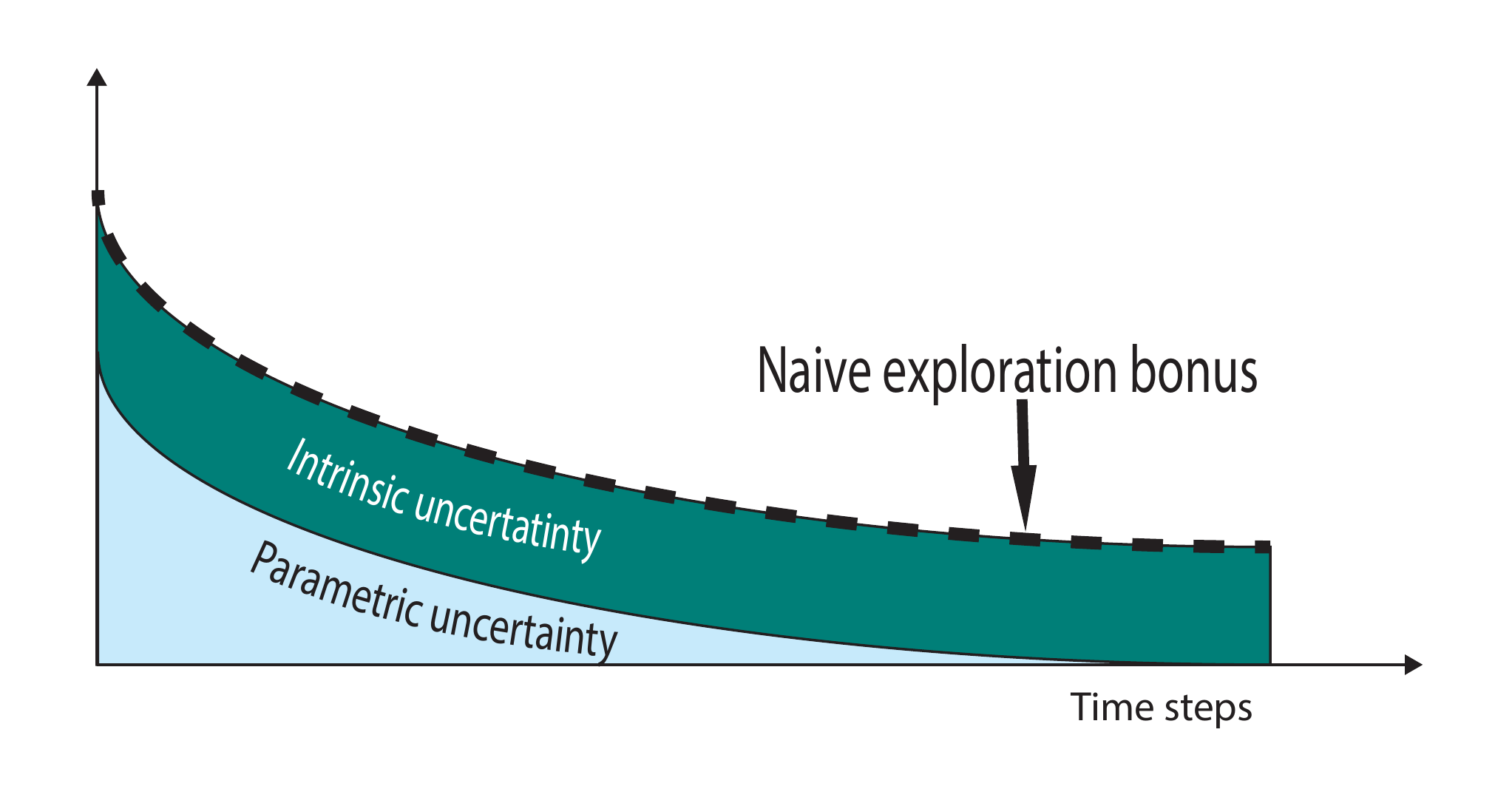}
            \caption{Naive exploration bonus.}
            \label{fig:schedule_a}
        \end{subfigure}
        
        \begin{subfigure}{0.42\textwidth}
        \includegraphics[width=1\textwidth]{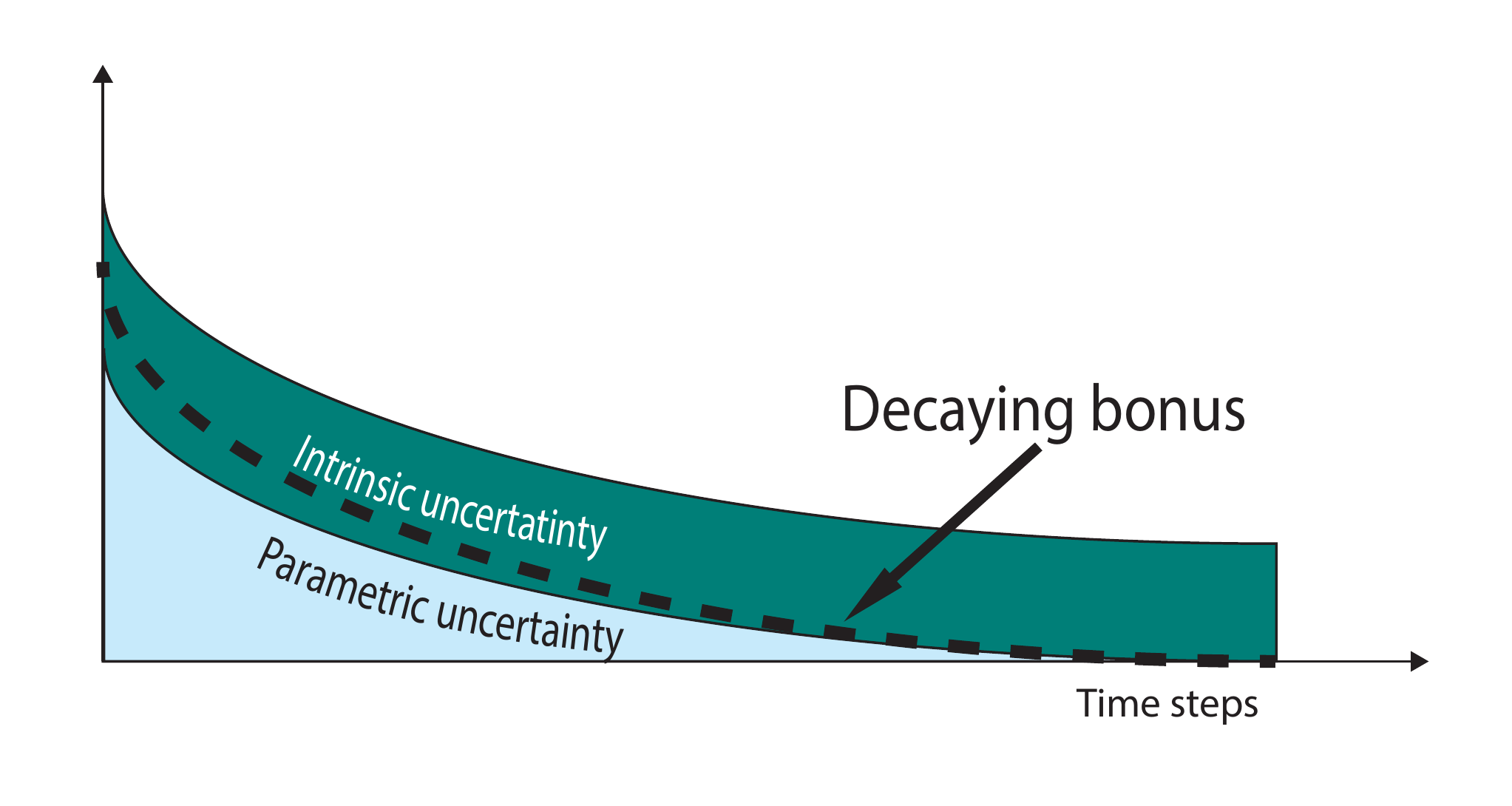}
            \caption{Decaying exploration bonus.}
            \label{fig:schedule_b}
        \end{subfigure}
        \caption{Exploration in the face of intrinsic and parametric uncertainties.}
    \end{figure}

\begin{figure}
        \centering
        \includegraphics[width=0.42\textwidth]{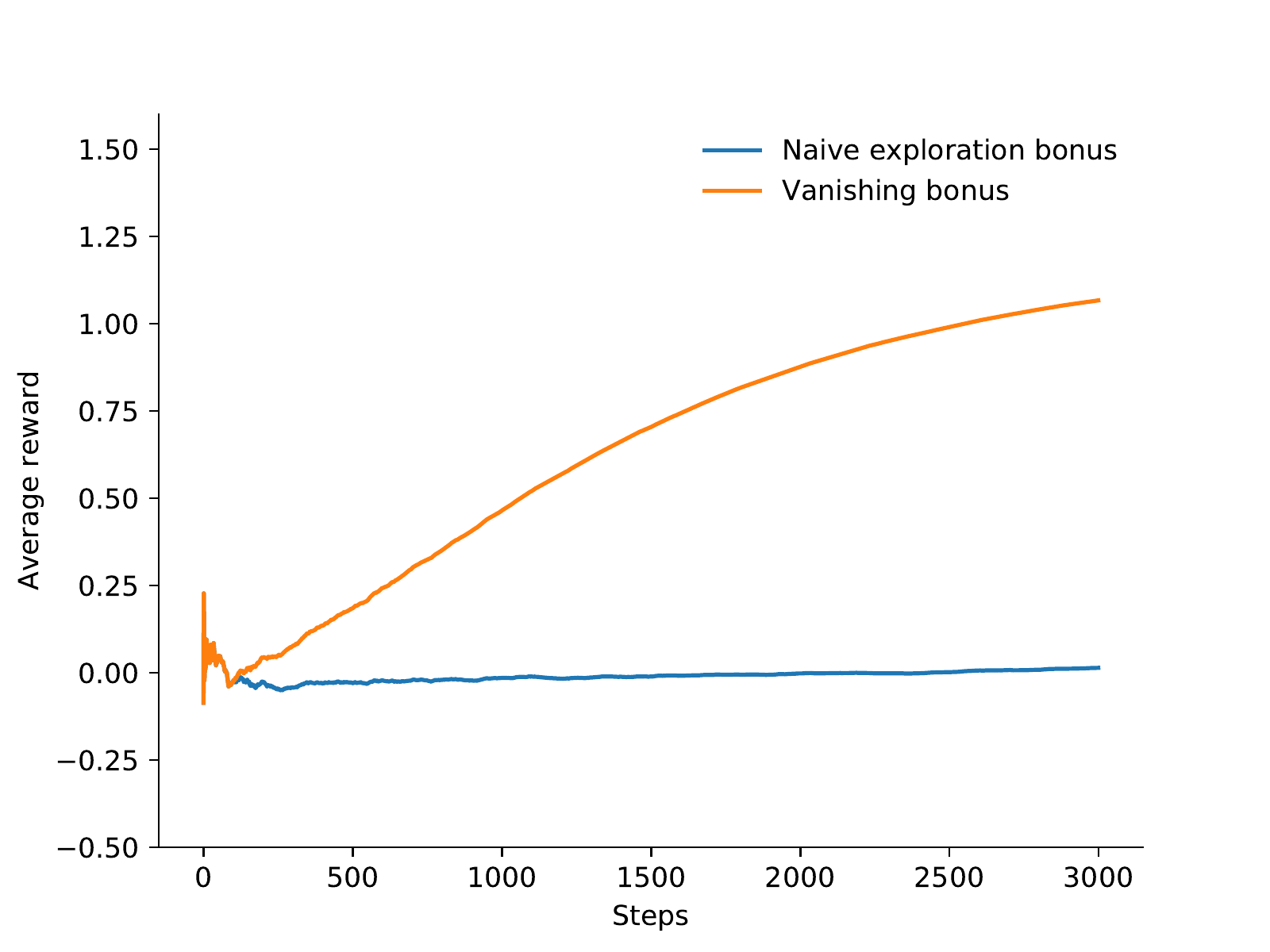}
            \caption{Performance of naive exploration and decaying exploration bonus in the counter example.}
            \label{fig:bandits_naive_vanishing}
    \end{figure}

    Figure~\ref{fig:bandits_naive_vanishing} shows that naive exploration bonus fails. Figure~\ref{fig:schedule_a}  illustrates the reason for the failure of naive exploration bonus. The estimated QR distribution is a mixture of parametric and intrinsic uncertainties. Recall, as learning progresses the parametric uncertainty vanishes and the intrinsic uncertainty stays (Figure~\ref{fig:schedule_b}). Therefore, this naive exploration bonus will tend to be biased towards intrinsic variation, which hurts performance. Note that the best arm has a low intrinsic variation. It is not chosen since its exploration bonus term is much smaller than the other arms as parametric uncertainty vanishes in all arms.
    
    The major obstacle in using the estimated distribution by QR for exploration is the composition of parametric and intrinsic uncertainties, whose variance is measured by the term $\sigma^2_k$ in (\ref{eq:naive_bandit}). To suppress the intrinsic uncertainty, we propose a decaying schedule in the form of a multiplier to $\sigma^2_k$:
    \begin{equation} \label{eq:schdule_with_var}
        a = \arg \max_k \bar{\mu}_k + c_t \bar{\sigma}_k 
    \end{equation}
    
    Figure~\ref{fig:schedule_b} depicts the exploration bonus resulting from the application of decaying schedule. From the classical QR theory \cite{koenker2005quantile}, it is known that the parametric uncertainty decays at the following rate:
    \begin{equation} \label{eq:schedule}
       c_t = c \sqrt{\frac{\log t}{t}}
    \end{equation}
    where $c$ is a constant factor. 

We apply this new schedule to the counter example where the naive solution fails. As shown in Figure~\ref{fig:bandits_naive_vanishing}, this decaying schedule significantly outperforms the naive exploration bonus.



    \subsection{Assymetry and truncated variance}
    
    \begin{figure*}
    \centering
        \begin{subfigure}{0.45\textwidth}
            \includegraphics[width=1\textwidth]{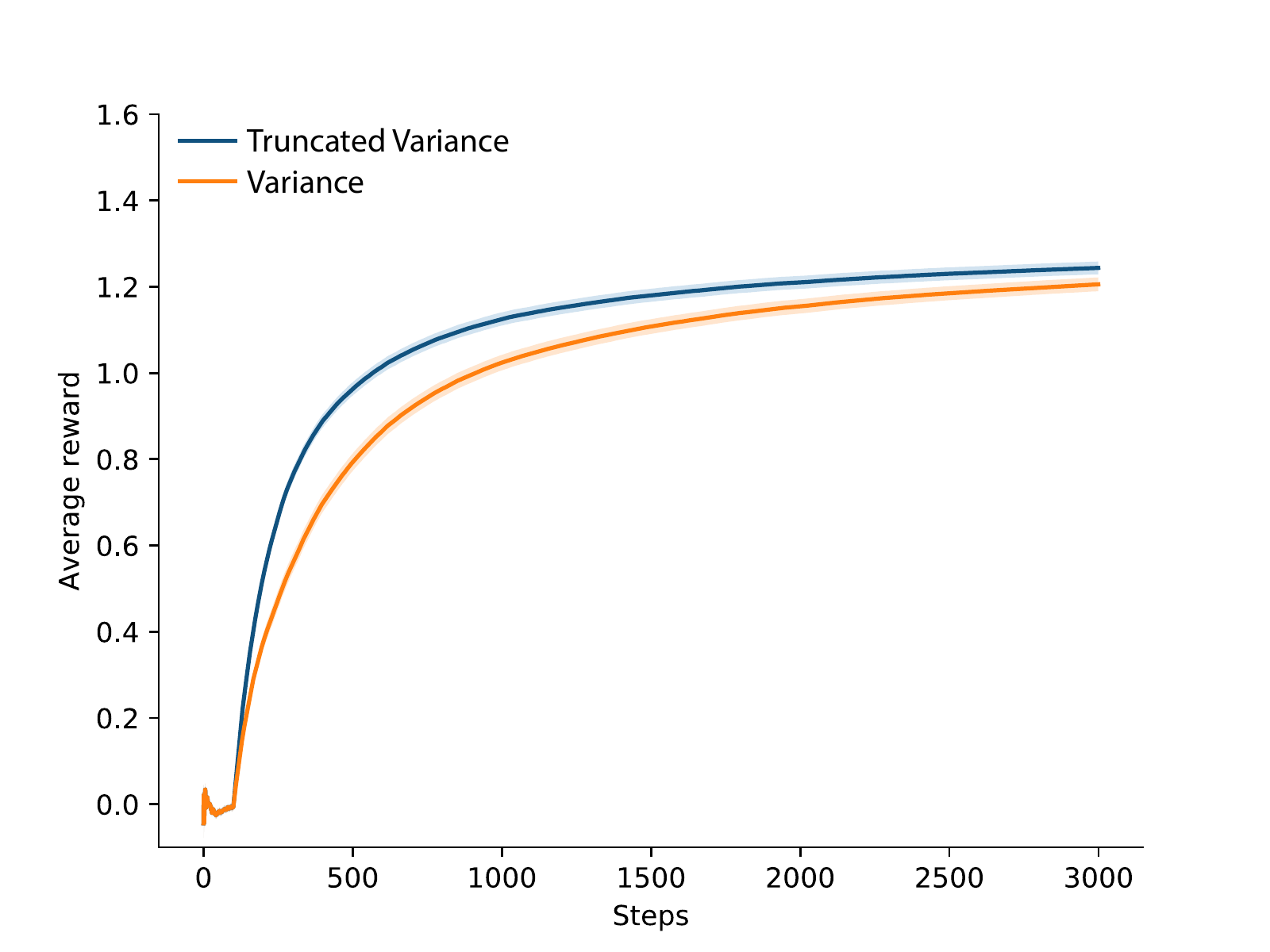}
            \caption{Environment with Symmetric distributions.}
        \end{subfigure}
        \begin{subfigure}{0.45\textwidth}
            \includegraphics[width=1\textwidth]{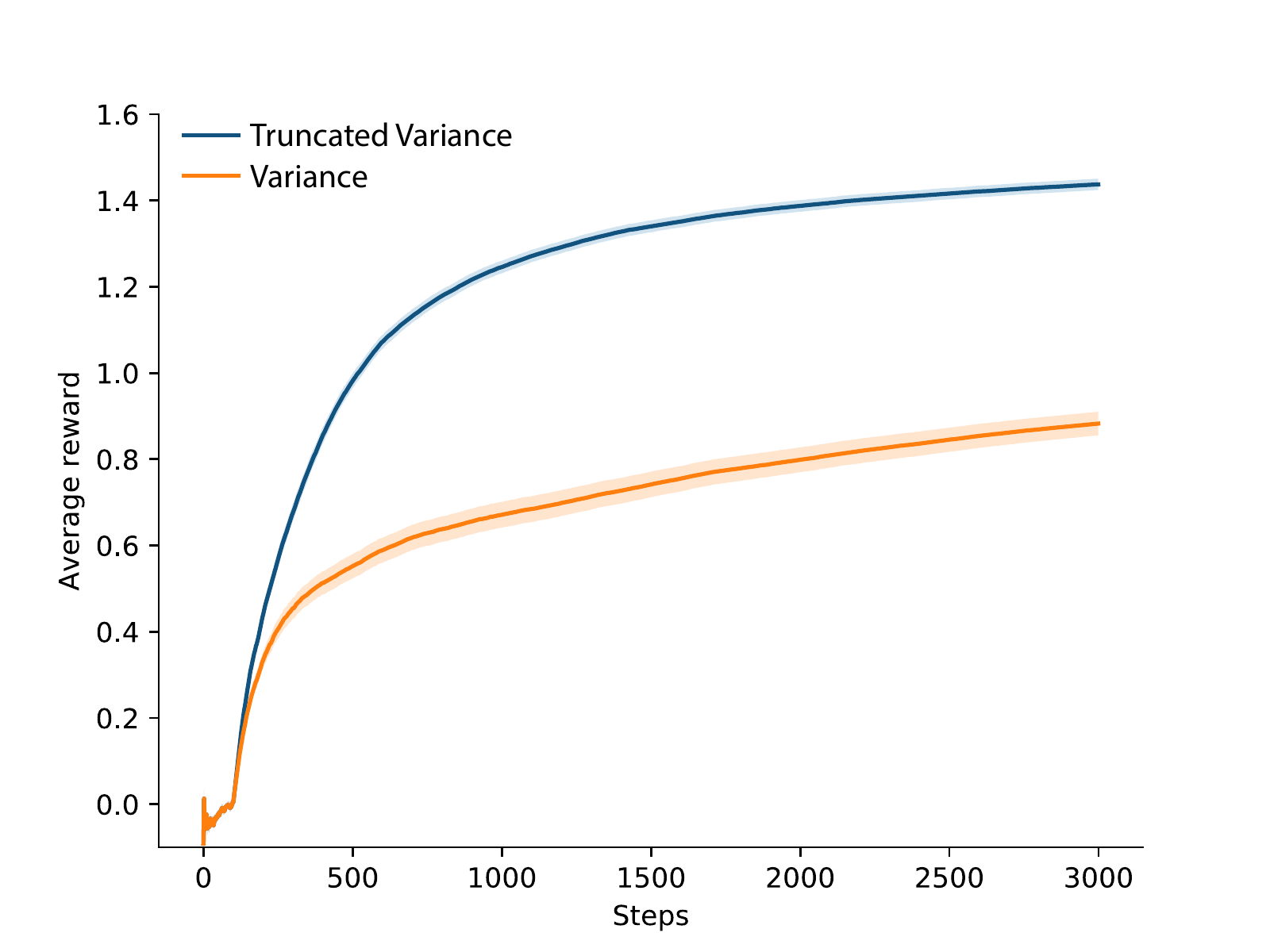}
            \caption{Environment with Asymmetric distributions.}
        \end{subfigure}
        \caption{Environments with symmetric and asymmetric rewards distributions.}
        \label{fig:bandits_asym}
        
    \end{figure*}
    
\begin{figure}
    \centering
        \includegraphics[width=0.45\textwidth]{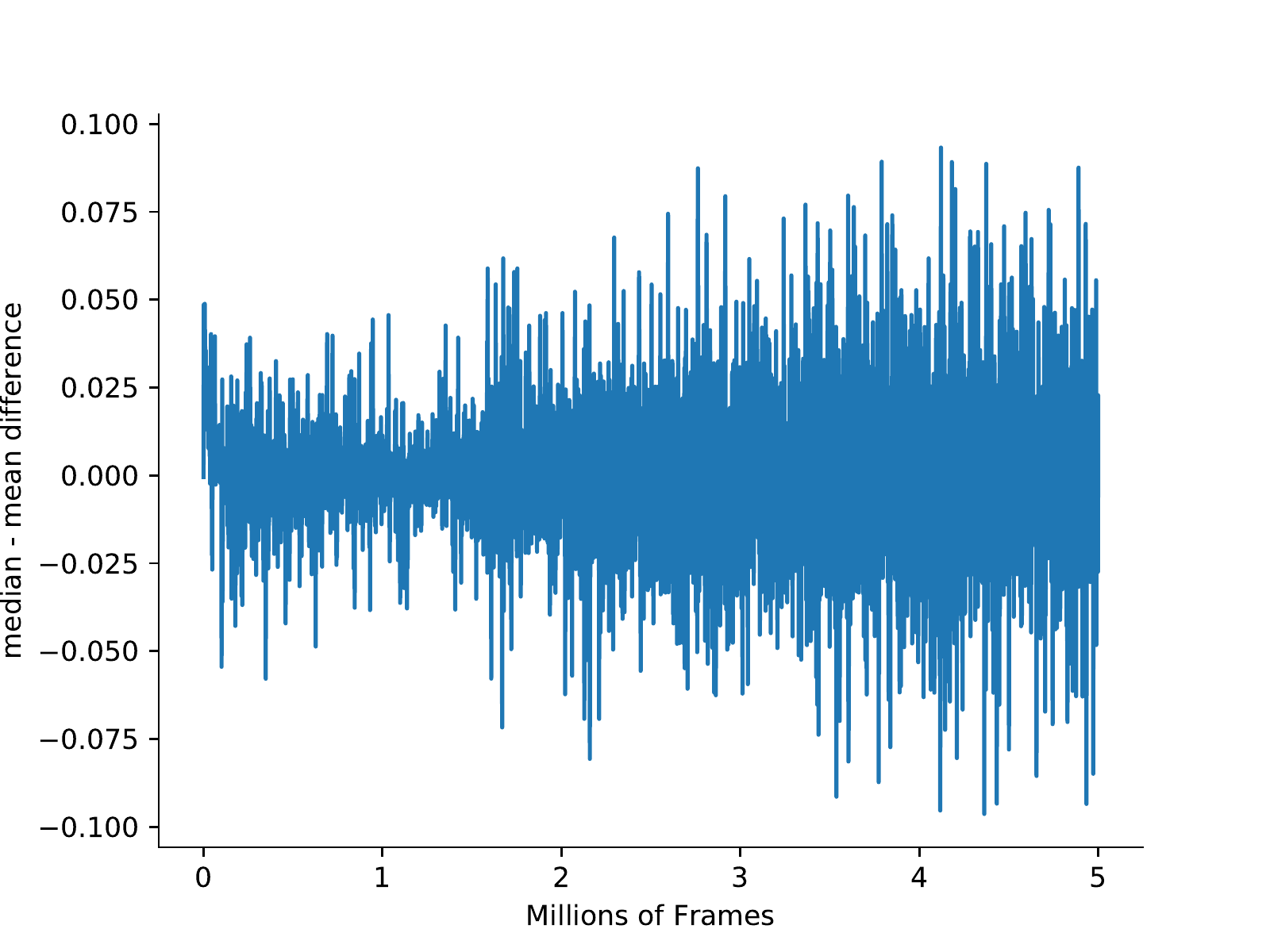}
        \caption{Pong. Empirical measure of the distribution learned for a single action obtained from QR-DQN-1 during training, showing very asymmetric.}
        \label{fig:pong_results_distr}
    \end{figure}

    QR has no restriction on the family of distributions it can represent. In fact, the learned distribution can be {\it asymmetric}, defined by mean $\ne$ median. From Figure~\ref{fig:pong_results_distr} it can be seen that the distribution estimated by QR-DQN-1 is mostly asymmetric. At the end of training, agent achieved nearly maximum score. Hence, the distributions correspond to the near-optimal policy, but they are not symmetric. 

    
    For the sake of the argument, consider a simple decomposition of the variance of the QR's estimated distribution into the two terms: the \textit{Right Truncated} and \textit{Left Truncated} variances \footnote{Note: Right truncation means dropping \textit{left} part of the distribution with respect to the mean}:
    \begin{equation*} \label{eq:variance_decomp}
        \begin{split}
        \sigma^2 = & \frac{1}{N}\sum_{i=1}^N(\bar{\theta} - \theta_i)^2 \\
        = & \frac{2}{N} \sum_{i=1}^\frac{N}{2}(\bar{\theta} - \theta_i)^2 + 
        \frac{2}{N} \sum_{i={\frac{N}{2}} + 1}^N(\bar{\theta} - \theta_i)^2 \\
        = & \sigma^2_{rt} + \sigma^2_{lt}, 
        \end{split}
    \end{equation*}
where $ \sigma^2_{rt}$ is the Right Truncated Variance and $ \sigma^2_{lt}$ is the right. 
To simplify notation we assume $N$ is an even number here. 
The Right Truncated Variance tells about lower tail variability and the Left Truncated Variance tells about upper tail variability.
In general, the two variances are not equal. \footnote{Consider discrete empirical distribution with support $\{-1, 0, 2\}$ and probability atoms $\{\frac{1}{3}, \frac{1}{3}, \frac{1}{3}\}$.} 
If the distribution is symmetric, then the two are the same.  
    
The Truncated Variance is equivalent to the Tail Conditional Variance (TCV):
    \begin{equation}
        TCV_x(\theta) = Var(\theta - \bar{\theta}| \theta > x)
    \end{equation}
    defined in \cite{valdez2005tail}. For instantiating optimism in the face of uncertainty, the upper tail variability is more relevant than the lower tail, especially if the estimated distribution is asymmetric \cite{valdez2005tail}. Intuitively speaking, $\sigma^2_{lt}$ is more optimistic. $\sigma^2_{lt}$ is biased towards positive rewards. To increase stability, we use the left truncated measure of the variability, $\sigma^2_+$, based on the median rather than the mean due to its well-known statistical robustness \cite{huber2011robust,hampel2011robust}:


    \begin{equation} \label{eq:sigma-plus}
        \sigma^2_+ = \frac{1}{2N} \sum_{i=\frac{N}{2}}^N(\theta_{\frac{N}{2}} - \theta_i)^2 
    \end{equation}
    where $\theta_i$'s are $\frac{i}{N}$-th quantiles.
    By combining decaying schedule from (\ref{eq:schedule}) with $\sigma^2_+$ from (\ref{eq:sigma-plus}) we obtain a new exploration bonus for picking an action, which we call Decaying Left Truncated Variance (DLTV). 

    In order to empirically validate our new approach we employ a multi-armed bandits environment with asymmetrically distributed rewards. In each run the means of arms $\{\mu_k\}_k$ are drawn from standard normal distribution. The best arm's reward follow $\mu_k + E[LogNormal(0,1)] - LogNormal(0,1)$. Other arms rewards follow $\mu_k + LogNormal(0,1) - E[LogNormal(0,1)]$. We compare the performance of both exploration methods in another, symmetric environment with rewards following the normal distribution centered at corresponding means (same as the asymmetric environment) with unit variance. 
    
    The results are presented in Figure~\ref{fig:bandits_asym}. With asymmetric reward distributions, the truncated variance exploration bonus significantly outperforms the naive variance exploration bonus. In addition, the performance of truncated variance is slightly better in the symmetric case.
    
    

    \subsection{DLTV for Deep RL}
    
    So far, we introduced the decaying schedule to control the parametric part of the composite uncertainty. Additionally, we introduced a truncated variance to improve performance in environments with asymmetric distributions. These ideas generalize in a straightforward fashion to the Deep RL setting. Algorithm \ref{alg:DLTV_deeprl} outlines DLTV for Deep RL. Action selection step in line 2 of Algorithm~\ref{alg:DLTV_deeprl} uses exploration bonus in the form of $\sigma^2_+$ defined in (\ref{eq:sigma-plus}) and schedule $c_t$ defined in (\ref{eq:schedule}). 
    
    \begin{algorithm}
      \caption{DLTV for Deep RL} \label{alg:DLTV_deeprl}
      \begin{algorithmic}[1]
        \REQUIRE $w, w^-, (x, a, r, x'), \gamma \in [0, 1)$ \COMMENT{network weights, sampled transition, discount factor}
        \STATE $Q(x', a') = \sum_j q_j \theta_j(x',a'; w^-)$
        \STATE $a^* = \arg \max_{a'}(Q(x, a') + c_t \sqrt{\sigma^2_+})$
        \STATE $\mathcal{T}\theta_j = r + \gamma \theta_j(x', a^*; w^-)$
        \STATE $L(w) = \sum_i \frac{1}{N} \sum_j[\rho_{\hat{\tau}_i}(\mathcal{T}\theta_j - \theta_i(x, a ; w))]$
        \STATE $w' = \arg \min_w L(w)$
        \ENSURE $w'$ \COMMENT{Updated weights of $\theta$}
      \end{algorithmic}
    \end{algorithm}
    
\begin{figure}
    \centering
        \includegraphics[width=0.47\textwidth]{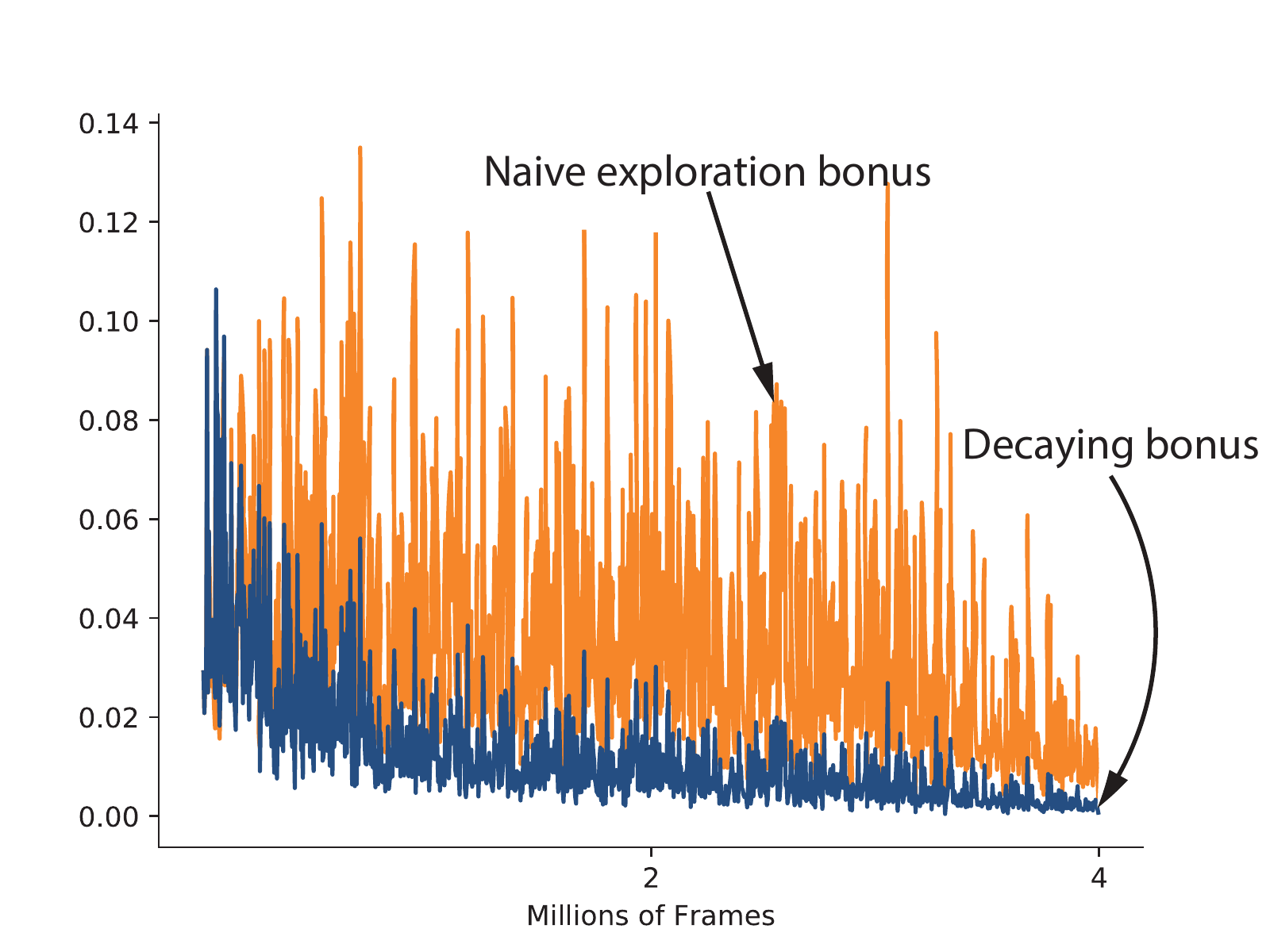}
        \caption{The naive exploration bonus and decaying bonus used for DLTV in Pong.}
        \label{fig:pong_results_exploration}
    \end{figure}

    Figure \ref{fig:pong_results_exploration} presents naive and decaying exploration bonus term from DLTV of QR-DQN during training in Atari Pong. Comparison of Figure~\ref{fig:pong_results_exploration} to Figure~\ref{fig:schedule_b} reveals the similarity in the behavior of the naive exploration bonus and the decaying exploration bonus.
    This shows what the raw variance looks like in Atari 2600 game and the suppressed intrinsic uncertainty leading to a decaying bonus as illustrated in Figure~\ref{fig:schedule_b}.
\section{Atari 2600 Experiments} \label{sec:experiments}
    
    
  \begin{figure}
        \centering
            \includegraphics[width=0.47\textwidth]{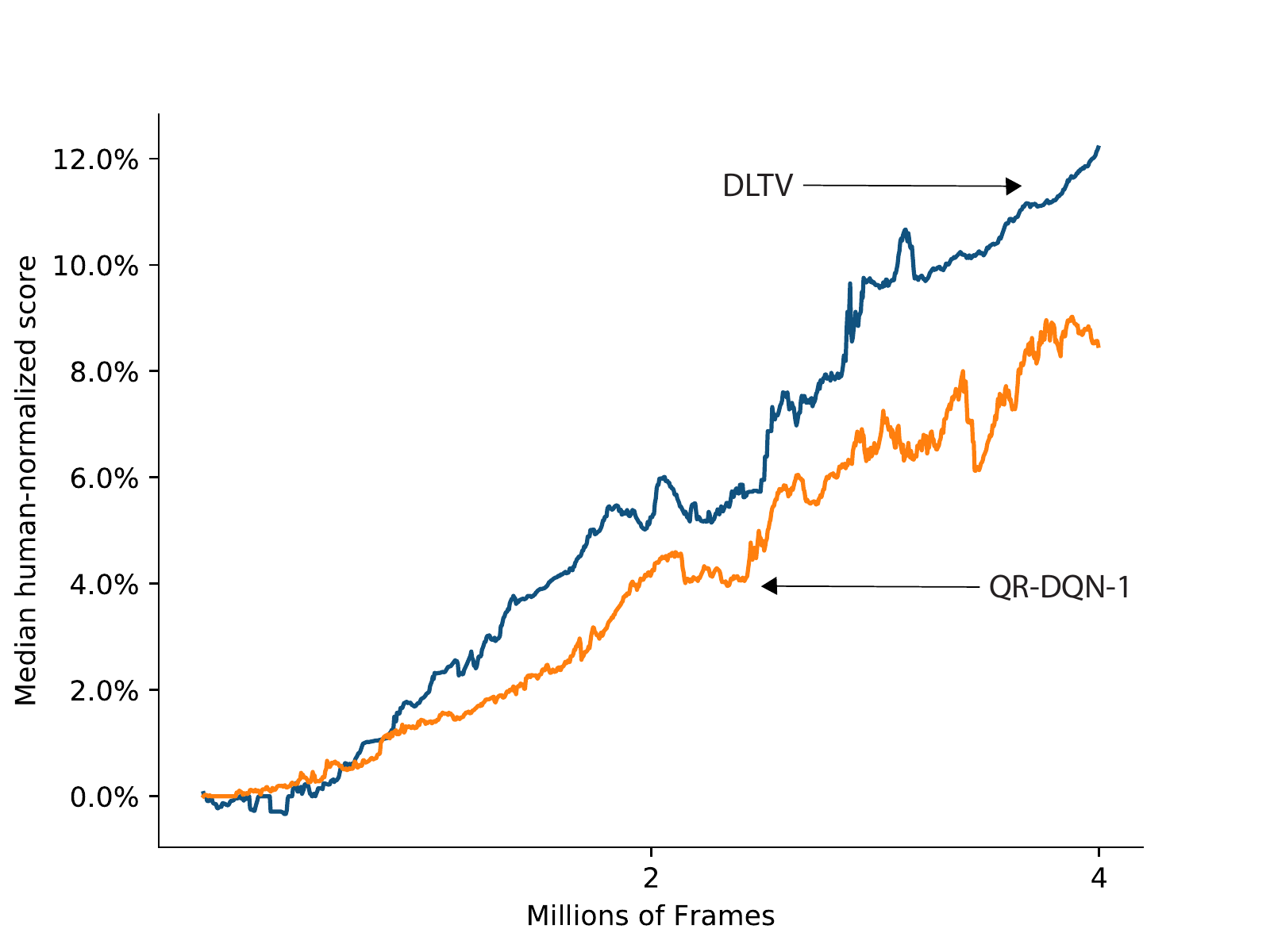}
        \caption{Median human-normalized performance across 49 games.}
        \label{fig:atari_results_median}
    \end{figure}

    We evaluated DLTV on the set of 49 Atari games initially proposed by \cite{mnih2015human}. Algorithms were evaluated on 40 million frames\footnote{Equivalently, 10 million agent steps.} 3 runs per game. The summary of the results is presented in Figure \ref{fig:atari_results}. Our approach achieved 483 \% average gain in cumulative rewards \footnote{The cumulative reward is a suitable performance measure for our experiments, since none of the learning curves exhibit plummeting behaviour. Plummeting is characterized by abrupt degradation of performance. In such cases the learning curve drops to the minimum and stays their indefinitely. A more detailed discussion of this point is presented in \cite{machado2017revisiting}.} over QR-DQN-1. Notably the performance gain is obtained in hard games such as Venture, PrivateEye, Montezuma Revenge and Seaquest. The median of human normalized performance reported in Figure~\ref{fig:atari_results_median} shows a significant improvement of DLTV over QR-DQN-1. We present learning curves for all 49 games in the Appendix.

    \begin{figure}
        \centering
            \includegraphics[width=0.47\textwidth]{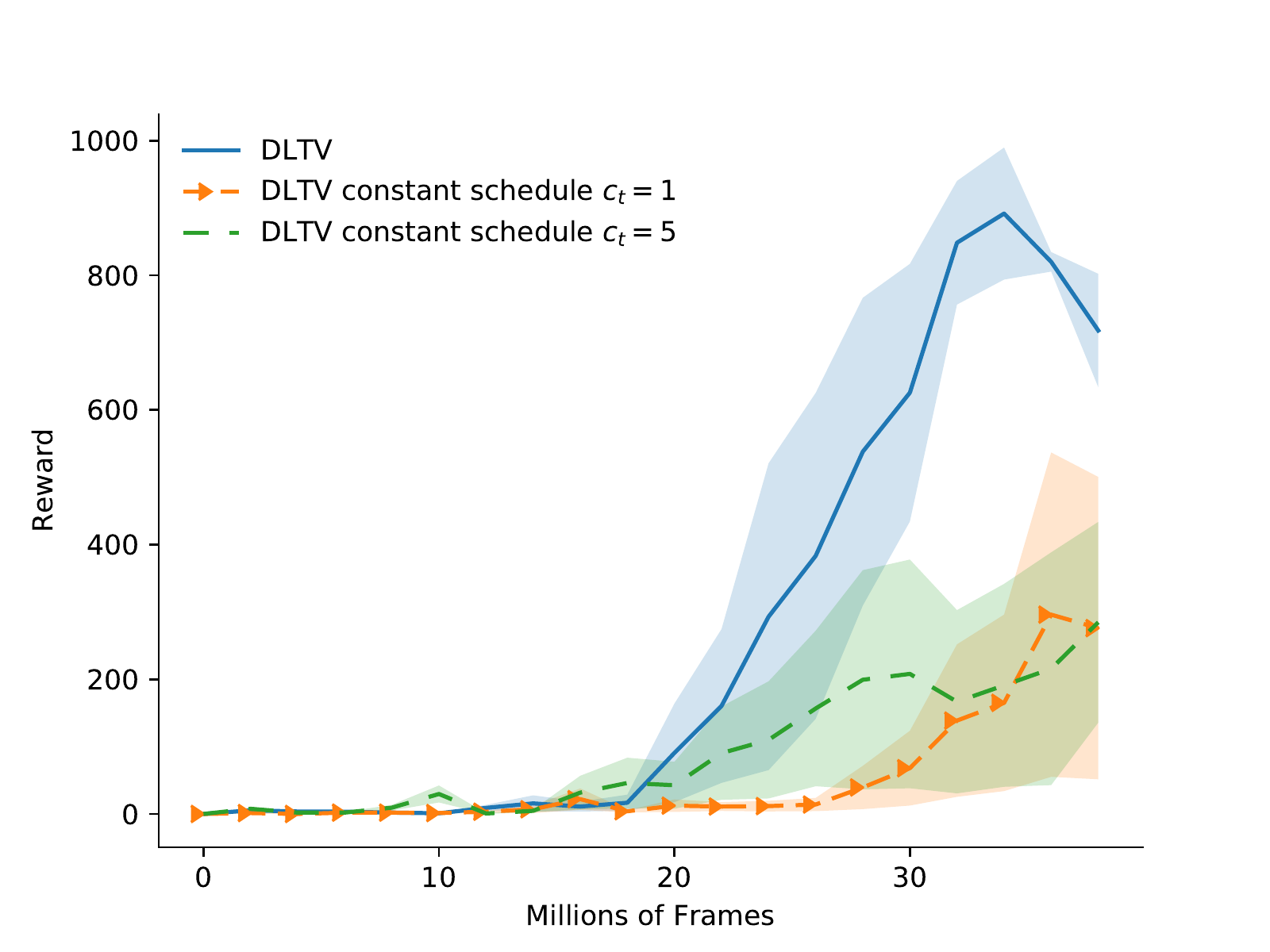}
        \caption{Online training curves for DLTV (with decaying schedule and with constant schedule) on the game of Venture.}
        \label{fig:venture_const}
    \end{figure}

    The architecture of the network follows \cite{dabney2017distributional}. For our experiments we chose the Huber loss with $\kappa=1$ \footnote{QR-DQN with $\kappa=1$ is denoted as QR-DQN-1} in the work by \cite{dabney2017distributional} due to its smoothness compared to $L1$ loss of QR-DQN-0. (Smoothness is better suited for gradient descent methods). We followed closely \cite{dabney2017distributional} in setting the hyper parameters, except for the learning rate of the Adam optimizer which we set to $\alpha=0.0001$.
    
    The most significant distinction of our DLTV is the way the exploration is performed. \emph{As opposed to QR-DQN there is no epsilon greedy exploration schedule in DLTV.} The exploration is performed via the $\sigma^2_+$ term only (line 2 of Algorithm 1). 
    
    An important hyper parameter which is introduced by DLTV is the schedule, i.e. the sequence of multipliers for $\sigma^2_+$,  $\{c_t\}_t$. In our experiments we used the following schedule $c_t = 50 \sqrt{\frac{\log t}{t}}$.
    


     We studied the effect of the schedule in the Atari 2600 game Venture. Figure \ref{fig:venture_const} show that constant schedule for DLTV significantly degenerates the performance. These empirical results show that the decaying schedule in DLTV is very important. 


\section{CARLA Experiments}
 A particularly interesting application of the (Distributional) RL approach is driving safety. 
There has been quite a converge of interests in using RL for autonomous driving, e.g.,  see \cite{sakib_iros,deeptraffic_mit,cc_lane,yao_mcts_uda}.
In the classical RL setting the agent only cares about the mean. In Distributional RL the estimate of the whole distribution allows for the construction of the risk-sensitive policies. For that reason we further validate DLTV in CARLA environment which is a 3D self driving simulator.
    
    \subsection{Sample efficiency}
\begin{figure}[t]
    \centering
        \includegraphics[width=0.47\textwidth]{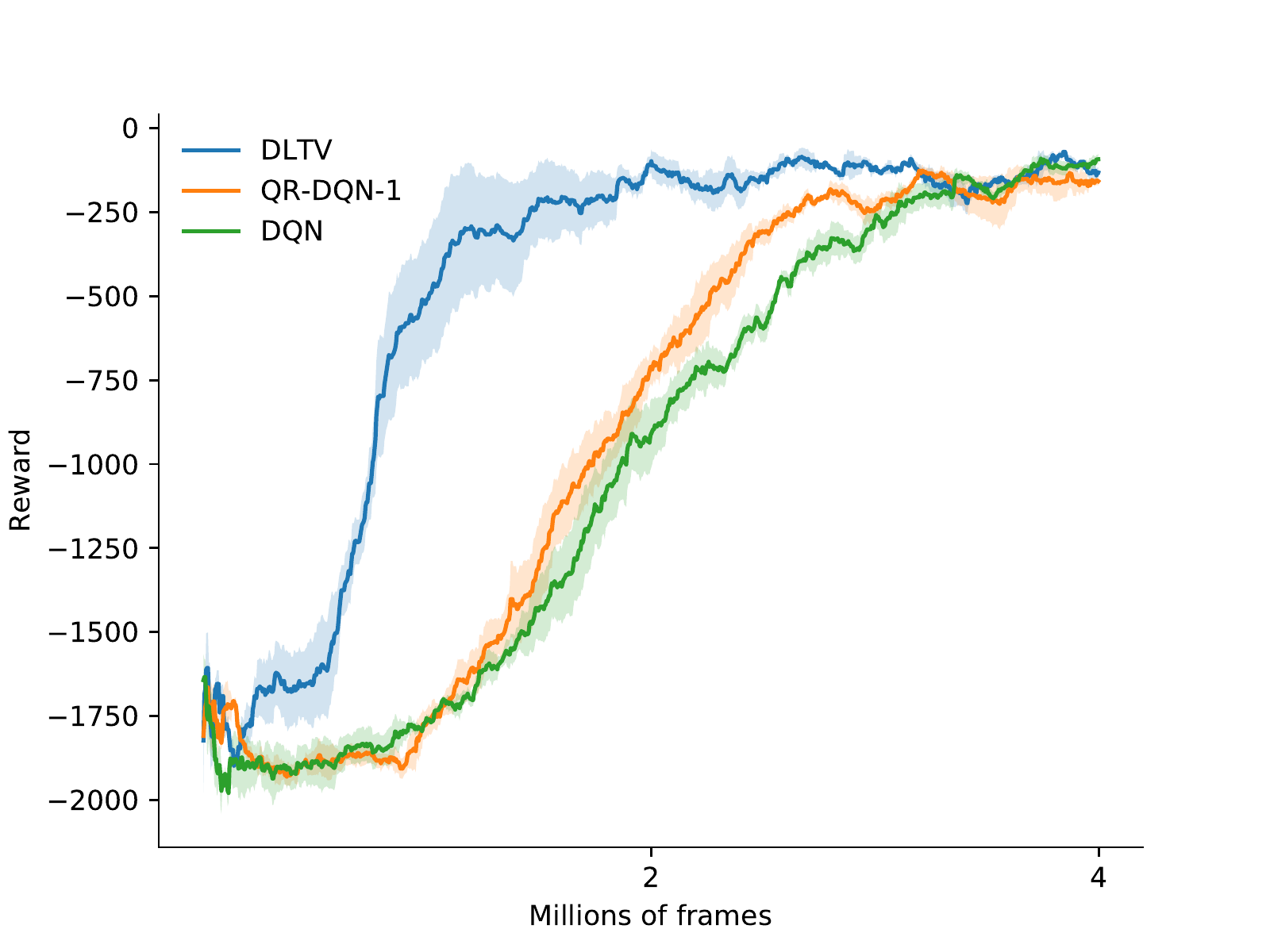}
        \caption{Naive exploration bonus and decaying bonus (as used in DLTV) for CARLA.
	DLTV learns significantly faster than DQN and QR-DQN, achieving higher rewards for safety driving.  
}
        \label{fig:carla_lc}
    \end{figure}

    It should be noted that CARLA is a more visually complex environment than Atari 2600, since it is based on a modern Unreal Engine 4 with realistic physics and visual effects.
    For the purpose of this study we picked the task in which the ego car has to reach a goal position following predefined paths. In each episode the start and goal positions are sampled uniformly from a predefined set of locations (around 20). We conducted our experiments in Town 2.
    We simplified the reward signal provided in the original paper \cite{dosovitskiy2017carla}. We assign reward of $-1.0$ for any type of infraction and a a small positive reward for travelling in the correct direction without any infractions, i.e. $0.001 (distance_{t} - distance_{t+1})$. The infractions we consider are: collisions with cars, collisions with humans, collisions with static objects, driving on the opposite lane and driving on a sidewalk. The continuous action space was discretized in a coarse grain fashion. We defined 7 actions: 6 actions for going in different directions using fixed values for steering angle and throttle and a no op action.
    The training learning curves are presented in Figure \ref{fig:carla_lc}. DLTV significantly outperforms QR-DQN-1 and DQN. Interestingly QR-DQN-1 performs on par with DQN. 
    
    \subsection{Driving Safety}
    A byproduct of Distributional RL is the estimated distribution of $Q(s, a)$. The access to this density allows for different approaches to control. For example \citet{morimura2012parametric} derive risk-sensitive policies based on the quantiles rather than the mean. The reasoning behind such approach is to view quantile as a risk metric.
    For instance, one particularly interesting risk metric is Value-at-Risk (VaR) which has been in use for a few decades in Financial Industry \cite{philippe2001value}.
    \citet{artzner1999coherent} define $VaR_\alpha(X)$ as $Prob(X \le -VaR_\alpha(X)) = 1- \alpha$, that is $VaR_\alpha(X) = (1 -\alpha) th$ quantile of $X$.

    It might be easier to understand the idea behind VaR in financial setting. Consider two investments: first investment will lose 1 dollar of its value or more with 10\% probability ($VaR_{10\%} = 1$) and second investment will lose 2 dollars or more of its value with 5 percent probability ($VaR_{10\%} = 2$). Second investment is riskier than the first one, that is a risk-sensitive investor will pick an investment with the higher VaR.
    This same reasoning applies directly to RL setting. Here, instead of investments we deal with actions. risk-sensitive policy will pick the action that has highest VaR. For instance \citet{morimura2012parametric} showed in a simple environment of Cliff Walk the policy maximizing low quantiles yields paths further away from the dangerous cliff.

    Risk-sensitive policies are not only applicable to toy domains. In fact risk sensitive policies is a very important research question in self-driving. In that respect CARLA is a non trivial domain where risk-sensitive policies can be thoroughly tested. In \cite{dosovitskiy2017carla} authors introduce simple safety performance metric such as average distance travelled between infractions. In addition to this metric we also consider the collision impact. This metric allows one to differentiate policies with the same average distance between infractions. Given the impact is not avoidable, a good policy should minimize the impact.

    \begin{table}[t] 
        \begin{tabular}{lll}
        \textbf{Average distance} & $\boldsymbol{VaR_{90\%}}$ or $\boldsymbol{q_{0.1}}$ & \textbf{Mean}    \\
        \textbf{between infractions} & & \\
        \hline
        Opposite lane            & \textbf{4.55}    & 1.35    \\
        Sidewalk                 & \textbf{None}    & None    \\
        Collision-static         & \textbf{None}    & 3.54    \\
        Collision-car            & 0.70    & \textbf{1.53}    \\
        Collision-pedestrian     & \textbf{52.33}   & 16.41   \\
                                 &         &         \\          
        \textbf{Average collision impact} &         &         \\
        \hline
        Collision-static         & \textbf{None}    & 509.81  \\
        Collision-car            & \textbf{497.22}  & 1078.76 \\
        Collision-pedestrian     & 40.79   & \textbf{40.70}   \\
                                 &         &         \\  
        \hline
        Distance, km             & \textbf{104.69}  & 98.66   \\
        \# of evaluation episodes       & 1000 & 1000
        \end{tabular}
        \caption{Safety performance in CARLA. 
	We compared decision making using mean and quantile, both are according to the model trained by DLTV. 
         Recall that DLTV learns a distribution of state-action values, represented by a set of quantile values. 
         On the middle column is selecting actions using a low quantile for the state-action value function, $q_{0.1}$, which is more conservative than the mean. In 1000 episodes, the total distance driven is $104.69$km, and driving on the opposite lane every $4.55$ km. Using the mean for action selection, the total distance driven is $98.66$ km and on opposite lane every $1.35$ km.
Across all measures, using low quantile achieves better than using mean for action selection, except that collision rate with car is higher but the collision impact is lower.     
}
        \label{table:safety}
    \end{table}

    We trained our agent using DLTV approach and during evaluation we used risk-sensitive policy derived from $VaR(Q(s, a)_{90\%})$ instead of the usual mean. Interestingly, this approach does employ mean-centered RL at all. We benchmark this approach against the agent that uses mean for control.
    The safety results for the risk-sensitive and the mean agents are presented in Table \ref{table:safety}. It can be seen that risk-sensitive agent significantly improves safety performance across almost all metrics, except for collisions with cars. However, the impact of colliding with cars is twice lower for the risk-sensitive agent.

\section{Related Work}
    \citet{tang2018exploration} combined Bayesian parameter updates with distributional RL for efficient exploration. However, they demonstrated improvement in only simple domains. \citet{zhang2018quota} generated risk-seeking and risk-averse policies via distributional RL for exploration, making use of both optimism and pessimism of intrinsic uncertainty. To our best knowledge, we are the first to use the parametric uncertainty in the estimated distributions learned by distributional RL algorithms for exploration.

    For optimism in the face of uncertainty in deep RL setting, \citet{bellemare2016unifying} and \citet{ostrovski2017count} exploited a generative model to enable pseudo-count. \citet{tang2017exploration} combined task-specific features from an auto-encoder with similarity hashing to count high dimensional states. \citet{chen2017ucb} used $Q$-ensemble to compute variance-based exploration bonus. \citet{o2017uncertainty} used uncertainty Bellman equation to propagate the uncertainty through time steps. Most of those approaches bring in non-negligible computation overhead. In contrast, our DLTV achieves this optimism via distributional RL (QR-DQN in particular) and requires very little extra computation.

\section{Conclusions}
    Recent advancements in distributional RL, not only established new theoretically sound principles but also achieved state-of-the-art performance in challenging high dimensional environments like Atari 2600. We take a step further by studying the learned distributions by QR-DQN, and discovered 
    the composite effect of intrinsic and parametric uncertainties is challenging for efficient exploration. In addition, the distribution estimated by distributional RL can be asymmetric.  
    We proposed a novel decaying scheduling to suppress the intrinsic uncertainty, and a truncated variance for calculating exploration bonus, resulting in a new exploration strategy for QR-DQN. Empirical results showed that the our method outperforms QR-DQN (with epsilon-greedy strategy) significantly in Atari 2600.
    Our method can be combined with other advancements in deep RL, e.g. Rainbow \cite{hessel2017rainbow}, to yield yet better results.


\clearpage

\bibliography{/home/hengshuai/reference}

\bibliographystyle{icml2019}
    \begin{figure*}
        \centering
        \includegraphics[width=0.99\textwidth]{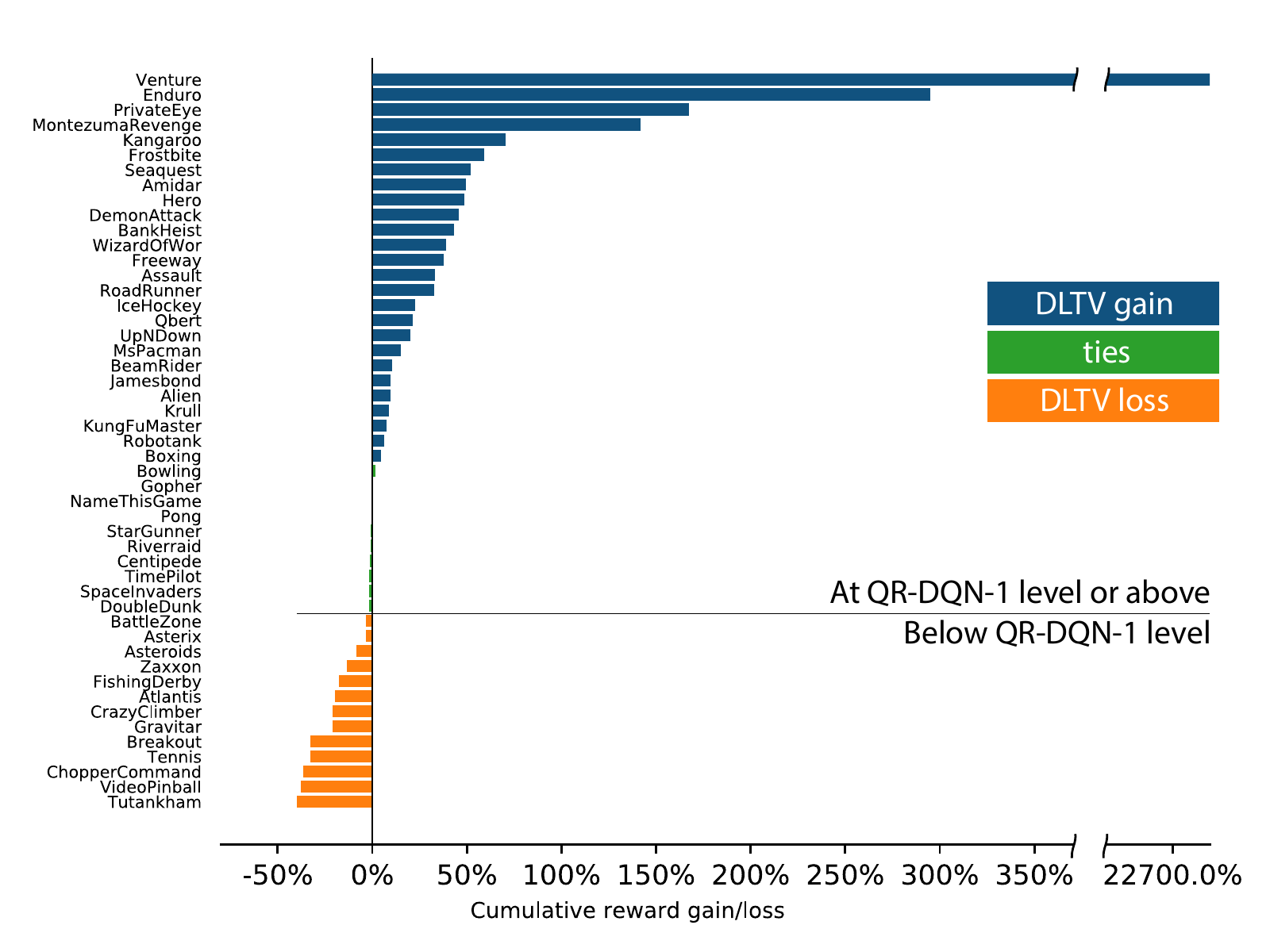}
        \caption{Cumulative rewards performance comparison of DLTV and QR-DQN-1. The bars represent relative gain/loss of DLTV over QR-DQN-1.}
        \label{fig:atari_results}
    \end{figure*}

\newpage
\newpage

\section*{Acknowledgement}
The correct author list for this paper is {\em Borislav Mavrin, Shangtong Zhang, Hengshuai Yao, Linglong Kong, Kaiwen Wu and Yaoliang Yu}. Due to time pressure, Shangtong's name was forgotten during submiting. If you cite this paper, please use this correct author list. The mistake was fixed in the arxiv version of this paper.

\appendix
\section{Performance Profiling on Atari Games}
Figure \ref{fig:atari_results} shows the performance of DLTV and QR-DQN on 49 Atari games, 
which is measured by cumulative rewards (normalized Area Under the Curve).

\end{document}